\documentclass[conference]{IEEEtran}
\IEEEoverridecommandlockouts
\usepackage{cite}
\usepackage{amsmath,amssymb,amsfonts}
\usepackage{algorithmic}
\usepackage{graphicx}
\usepackage{textcomp}
\usepackage{xcolor}

\usepackage{hyperref}

\def\BibTeX{{\rm B\kern-.05em{\sc i\kern-.025em b}\kern-.08em
    T\kern-.1667em\lower.7ex\hbox{E}\kern-.125emX}}
\begin{document}

\title{VEMOCLAP: A video emotion classification web application 
\thanks{This work has been funded by National Funds through the Portuguese funding agency, FCT - Fundação para a Ciência e a Tecnologia, within project LA/P/0063/2020. Serkan Sulun received the support of fellowship FCT - Fundação para a Ciência e a Tecnologia with the fellowship code 2022.09594.BD.}
}

\author{\IEEEauthorblockN{Serkan Sulun}
\IEEEauthorblockA{\textit{INESC TEC}\\
Porto, Portugal \\
serkan.sulun@inesctec.pt}
\and
\IEEEauthorblockN{Paula Viana}
\IEEEauthorblockA{\textit{INESC TEC,}\\\textit{ISEP, Polytechnic of Porto}\\
Porto, Portugal \\
pmv@isep.ipp.pt\\ paula.viana@inesctec.pt}
\and
\IEEEauthorblockN{Matthew E. P. Davies}
\IEEEauthorblockA{\textit{Independent researcher} 
% \\
% New York, USA \\
% matthew.davies@siriusxm.com
}
}

\maketitle

\begin{abstract}
We introduce VEMOCLAP: Video EMOtion Classifier using Pretrained features, the first readily available and open-source web application that analyzes the emotional content of any user-provided video. We improve our previous work, which exploits open-source pretrained models that work on video frames and audio, and then efficiently fuse the resulting pretrained features using multi-head cross-attention. Our approach increases the state-of-the-art classification accuracy on the Ekman-6 video emotion dataset by $4.3\%$ and offers an online application for users to run our model on their own videos or YouTube videos. We invite the readers to try our application at \href{http://serkansulun.com/app}{serkansulun.com/app}.

\end{abstract}


\section{Introduction and related work}

We present a web application for classifying the emotion in any user-provided video. Hosted on Google Colab with free GPU runtime, it is accessible to users of all skill levels and requires only a few mouse clicks. Users can upload a video, link a YouTube video, or select from available sample videos. The application outputs predicted emotions and includes additional analyses such as automatic speech recognition (ASR), optical character recognition (OCR), face detection and facial expression classification, audio classification, and image captioning. 

We improve our previous work and train our revised model on video emotion classification. While we comprehensively explain our method, we invite the reader to view our previous work for an in-depth description \cite{eswa}. We train our models on the Ekman-6 video emotion dataset and achieve a new state-of-the-art classification accuracy. The Ekman-6 dataset, one of the largest publicly available video emotion datasets, contains $1637$ videos from YouTube and Flickr, each categorized into one of $6$ emotions: anger, disgust, joy, sadness, and surprise \cite{ekman6}. Our model not only surpasses previous state-of-the-art results with the original training and testing splits, but also benefits from data cleansing that improves the classifier used in our web application.

Our contributions are the following:

\begin{itemize}
    \item We improve the state-of-the-art classification accuracy on the Ekman-6 video emotion classification dataset by $4.3\%$.

    \item We inspect and clean the Ekman-6 dataset, providing a list of problematic samples to enhance the training of video emotion models.

    \item We introduce an open-source and readily available web application that allows users to analyze and classify emotions in any video by uploading it or providing a YouTube link.
\end{itemize}

Though Google Colab provides free GPUs, the CPU and GPU memory are limited to around $15$ GBs. We redesigned our previous work to reduce its computational complexity for seamless deployment on Google Colab. First, instead of using the entire video, we extracted and used a limited number of frames. We also replaced the transformer model with multi-headed cross-attention modules that efficiently handle the temporal dependencies between multimodal features \cite{transformer}.

Our model has a low memory footprint due to the use of open-source, readily available pretrained feature extractor models. These models run in inference mode, avoiding backpropagation and storing gradients. We claim that the features extracted by these pretrained models are highly relevant to a video's emotion. Therefore, we can fuse the extracted features efficiently and process them using shallow neural networks.

The pretrained models we used are as follows:

\textit{Face detector:} The face detection model from the Ultralytics group is based on YOLO (You Only Look Once) \cite{ultralytics}. YOLO is a real-time object detection algorithm that divides an image into a grid and predicts bounding boxes and class probabilities for each grid cell using convolutional neural networks (CNNs) \cite{yolo}.

\textit{Expression classifier:} Pakov has finetuned a Vision Transformer (ViT) on the FER-2013 (Facial Emotion Recognition) dataset \cite{expression_classifier}. The model takes a facial image and predicts the facial expression as angry, disgusted, fearful, happy, sad, surprised, or neutral.

\textit{CLIP and CLIPCap:} Contrastive Language-Image Pretraining (CLIP) is a prominent model for image understanding, trained with contrastive learning on a large dataset of images and captions from the internet \cite{clip}. CLIP generates encodings (features) from images, while CLIPCap uses these encodings to produce full captions \cite{clipcap}.

\textit{Optical Character Recognition (OCR):} The PaddleOCR model detects and extracts the text from video frames \cite{paddle}.

\textit{Automatic Speech Recognition (ASR):} OpenAI's Whisper model processes audio input, detects speech patterns, and outputs the text \cite{whisper}. It also identifies the language and translates non-English speech into English.

\textit{BEATs:} Bidirectional Encoder representation from Audio Transformers (BEATs) is an audio classification model that is composed of an acoustic tokenizer and a classifier that are trained iteratively \cite{beats}. 

\textit{Language identification and translation:} 
% Papariello \cite{language_detection} trained an XLM-RoBERTa model \cite{xlmroberta} on language identification datasets to classify the language of a given text. 
Papariello trained an XLM-RoBERTa model on language identification datasets to classify the language of a given text \cite{language_detection}. 
% Facebook's NLLB-200 (No Language Left Behind) \cite{translator} uses a Sparsely Gated Mixture of Experts model \cite{dynamic}, trained on internet-sourced text data, to translate between $200$ languages. 
Facebook's NLLB-200 (No Language Left Behind) uses a Sparsely Gated Mixture of Experts model, trained on internet-sourced text data, to translate between $200$ languages \cite{translator}. 
While the NLLB-200 model requires the source language to be specified, Papariello's model addresses this by identifying the source language automatically.

\textit{Spell correction:} The Symspell package provides several tools for spell checking \cite{symspell}. Among them, the word segmenter separates words in sentences where spaces are missing, which is particularly useful for post-processing OCR outputs that may lack spaces.
% Martynov \cite{spellcheck} trained a T5 transformer language model \cite{t5} on a dataset with synthetic spelling errors, enabling it to correct any English text.
Martynov trained a T5 transformer language model on a dataset with synthetic spelling errors, enabling it to correct any English text \cite{spellcheck}.

% \textit{Sentiment classifier:} The Cardiff NLP group \cite{sentiment} trained the RoBERTa language model \cite{roberta} on the TweetEval benchmark \cite{tweeteval}. 
\textit{Sentiment classifier:} The Cardiff NLP group trained the RoBERTa language model on the TweetEval benchmark \cite{sentiment}. 
This model can predict the sentiment of a given text as positive, negative, or neutral.

We made our web application, along with the codebase, trained classifier, extracted pretrained features, and data cleansing results, publicly available. The project main page can be found at \href{http://serkansulun.com/vemoclap}{serkansulun.com/vemoclap}, the web application is available at \href{http://serkansulun.com/app}{serkansulun.com/app}, and the extracted pretrained features are hosted at \href{https://zenodo.org/records/13624583}{zenodo.org/records/13624583}.

\section{Methodology}

\begin{figure}[htbp]
\centerline{\includegraphics[width=0.4\textwidth]{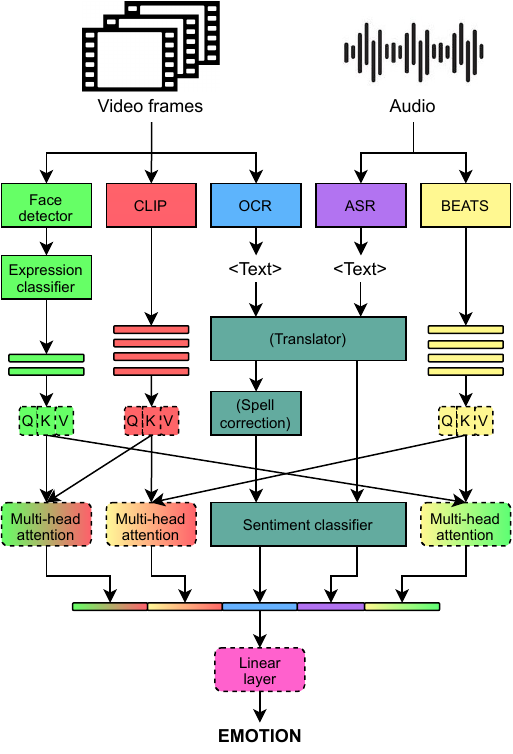}}
\caption{Video emotion classification pipeline. Blocks with rounded and dashed outlines represent trained modules. The models with parentheses are used conditionally. Other blocks are pretrained feature extractors and are used in inference mode. Q, K, and V represent query, key, and value projections.}
\label{fig:model}
\end{figure}

\subsection{Model}

Our video emotion classification pipeline can be seen in Figure \ref{fig:model} \cite{eswa}. It consists of frozen pretrained feature extractor models and trained modules for fusing and classifying the pretrained features into emotions.

\subsubsection{Pretrained feature extraction}

We initially extracted a fixed number $n$ of frames from each video, along with the entire audio, which was resampled at $16kHz$ and converted to mono. We then extracted relevant pretrained features in inference mode for all videos. Notably, for pretrained classifiers like the expression classifier, sentiment classifier, and BEATs, we use activations from the layers before the final classification rather than the final classifications. However, our web application also provides the final classifications for a more comprehensive video analysis.

The features from the facial expression classifier and CLIP are sequences of vectors, as their inputs are sequences of frames. If multiple faces are detected in a single video frame, we average the features of the two largest faces. Similarly, the BEATs model processes sequences of $3$-second audio chunks. We extracted $n$ audio chunks to match the number of video frames. While CLIP and BEATs produce $n$ output vectors, the facial expression classifier may produce fewer vectors, depending on whether faces are present in each frame. 

The ASR model generates a single block of text from the entire audio. If the source language is not English, it automatically translates the output text into English.

The OCR model generates blocks of text for each video frame. The language identifier processes each block. The text is translated into English if the identified language is not English. If the language is English, the text is passed through the word segmenter and then the spell corrector. The resulting text blocks from each frame are concatenated to form a single block of text.

The texts resulting from ASR and OCR are fed into the sentiment classifier separately. Since the sentiment classifier predicts a single label for any length of text, a single vector is extracted as the feature.

\subsubsection{Emotion classification}

After the feature extraction, for a single video, we are left with $n$ CLIP, $n$ BEATs, $k \leq n$ facial expression, $1$ OCR sentiment, and $1$ ASR sentiment features. Fusing these pretrained features presents multiple challenges. First, since they are extracted using different pretrained models, the lengths (dimensionalities) of the feature vectors are different. Secondly, when the feature vectors form a sequence, as in the case of facial expressions, CLIP, and BEATs, their temporal lengths can also be different. Finally, since they belong to different modalities, the content of these feature vectors can be vastly different.

% Fusing these pretrained features poses several challenges. First, the feature vectors have different dimensionalities because they come from various pretrained models. Second, for features forming sequences---such as those from facial expressions, CLIP, and BEATs---the temporal lengths may vary. Finally, due to the different modalities of these features, their content can be vastly different.
% To address these issues, we first performed min-max normalization on each feature using statistics extracted from the collection of pretrained features. The attention module first projects each input vector to a common dimensionality to handle differing dimensionalities.
% we processed each sequence of features—namely those from facial expressions, CLIP, and BEATs—individually using bidirectional gated recurrent units (Bi-GRUs) with matching input dimensionalities \cite{gru}. We used the Bi-GRUs as classifiers, extracting only the output from the final recurrence step. Thus, each Bi-GRU processes a sequence of vectors and produces a single vector. 
To address these issues, we first performed min-max normalization on each feature using statistics extracted from the collection of pretrained features. To handle differing dimensionalities, all sequential input features are first projected to queries, keys, and values with a common dimensionality $d$ \cite{transformer}. Next, the attention modules exploit correspondence between pairs of sequential features. The attention modules also include dropout and layer normalization and can handle a pair of sequences with different temporal lengths. As done in classification tasks, the attention outputs are averaged along the temporal dimension, yielding a single vector.
Since the OCR and ASR sentiment features are already single vectors, each modality is represented by a single vector after the attention modules. We then concatenated all five feature vectors along the channel dimension, resulting in a single vector representing the entire video. Finally, this vector was fed into a linear layer followed by a softmax layer, which outputs a probability for each emotion.

\subsubsection{Implementation details and hyperparameters}

We initially extracted video frames at $1$ frame per second, using $n=16$ video frames and audio chunks as input to our model. However, users can adjust the parameter $n$ during training or inference. During training, we selected the $n$ video frames and audio chunks from random locations for data augmentation. During testing, we extracted them at equidistant intervals to ensure comprehensive temporal representation. We reported classification performance using the provided training and testing splits, which included $819$ and $818$ videos, respectively.

We used cross-entropy loss, a batch size of $32$, a dropout rate of $0.5$, and Adam optimizer with a learning rate of $1e-5$ \cite{adam}. 
% Each Bi-GRU has only $1$ layer with a hidden layer dimensionality of $512$. During training, gradients of the Bi-GRUs are clipped at a norm of $1.0$, 
Attention modules have $4$ heads and a dimensionality of $512$. We used $10\%$ of the training split as the validation split and stopped training when validation accuracy started to drop. The model has around $11M$ trainable parameters.
% and we used a batch size of $32$. We employed early stopping when the test accuracy started to drop. The model has around $11M$ trainable parameters. We used cross-entropy loss with class weights inversely proportional to class frequencies in the dataset to address label imbalances.

% \begin{equation}
% \text{weight}_c = \frac{\frac{1}{\sum_{n=1}^N \mathbf{1}(y_n = c)}}{\sum_{i=1}^C \frac{1}{\sum_{n=1}^N \mathbf{1}(y_n = i)}}
% \end{equation}

% In the formula, \(c\) represents a specific class label, \(C\) is the total number of classes, and \(N\) is the total number of samples. The indicator function \(\mathbf{1}(y_n = c)\) is 1 if the class label \(y_n\) of the \(n\)-th sample equals \(c\), and 0 otherwise.

\subsection{Dataset cleaning}
\label{cleaning}
While we report classification results on the unedited Ekman-6 dataset, we cleaned it to train the model used in the web application. The Ekman-6 dataset was created by scraping the web for videos using search keywords that matched not only the categorized emotion but also related terms. We viewed each video to detect the problematic samples. 
After inspecting their file names, we identified the following problematic search keywords for each emotion class, which are \underline{underlined}.

\textbf{Anger}: A single person being \underline{annoy}ing, with no other person present to be annoyed or angry.

\textbf{Disgust}: Flashing lights or rapid camera movement, presumably to induce \underline{dizzy}ness or \underline{nausea}. It also includes videos related to \underline{boredom} and \underline{loathing}.

\textbf{Fear}: Counter-\underline{terror}ism, \underline{underwater} footage, 9/11 \underline{terror}ism attack aftermath, and suspect \underline{apprehension}.

\textbf{Joy}: \underline{Joy}ride (driving a car), the music "Ode to \underline{Joy}", and people named \underline{Joy}.

\textbf{Sadness}: \underline{pensive}

\textbf{Surprise}: \underline{distraction}, and people performing impressive feats labeled as \underline{astonishing}.

% TODO: train again with blacklist

We identified and removed $128$ and $130$ problematic videos from the training and testing splits, respectively. Using the cleaned data, the classification accuracy increased by $2.6\%$. However, we exclude this result from our comparison with the state-of-the-art because data cleaning alters the test split's content, affecting the comparison's fairness. For training the model used in our web application, we alphabetically sorted the video names for each category, used the first $95\%$ for training, and reserved the remaining $5\%$ for validation. We made the list of the problematic videos available as \textit{ekman\_blacklist.txt}

\subsection{Inference web application}

We developed an open-source web application for performing inference on user-provided videos. Hosted on Google Colab, it offers free GPU runtime. Users can upload their own videos, provide a YouTube link, or use sample videos provided within the application. The application is self-contained and ready to use, requiring no setup from the user. The process is streamlined into $5$ steps, with only $5$ mouse clicks needed to obtain the results. After connecting to a GPU runtime, users should follow these steps:

\textbf{Step 1:}
Automatically download and extract the codebase, and install the required Python libraries. This step takes approximately $2$ minutes.

\textbf{Step 2:}
Download and build the feature extractor models and the classifier model. As these models are deep neural networks, this step takes about $3$ minutes. Note that Steps 1 and 2 only need to be completed once, even if classifying multiple videos.

\textbf{Step 3:}
Select how to load the input video. The options are ``Sample video", ``YouTube link", and ``Upload video".

\textbf{Step 4:}
Depending on the choice from step 3, the user then selects the specific video. Steps 3 and 4 take only a few seconds to complete.

\textbf{Step 5:}
Extract the frames and audio from the input video, run the pretrained feature extractors, and finally run the emotion classifier. The outputs include text from automatic speech recognition (ASR) with its sentiment, text from optical character recognition (OCR) with its sentiment, and predictions from the BEATs audio classifier. Additionally, a sample frame is displayed showing detected faces with predicted emotional expressions, detected OCR boxes, and a caption generated by CLIPCap. Note that this sample frame is for demonstration purposes, while all $n$ frames are used for the final emotion classification. For a $60$-second video, this step takes approximately $30$ seconds.

\section{Experiments and results}

\begin{figure}[htbp]
\centerline{\includegraphics[width=0.45\textwidth]{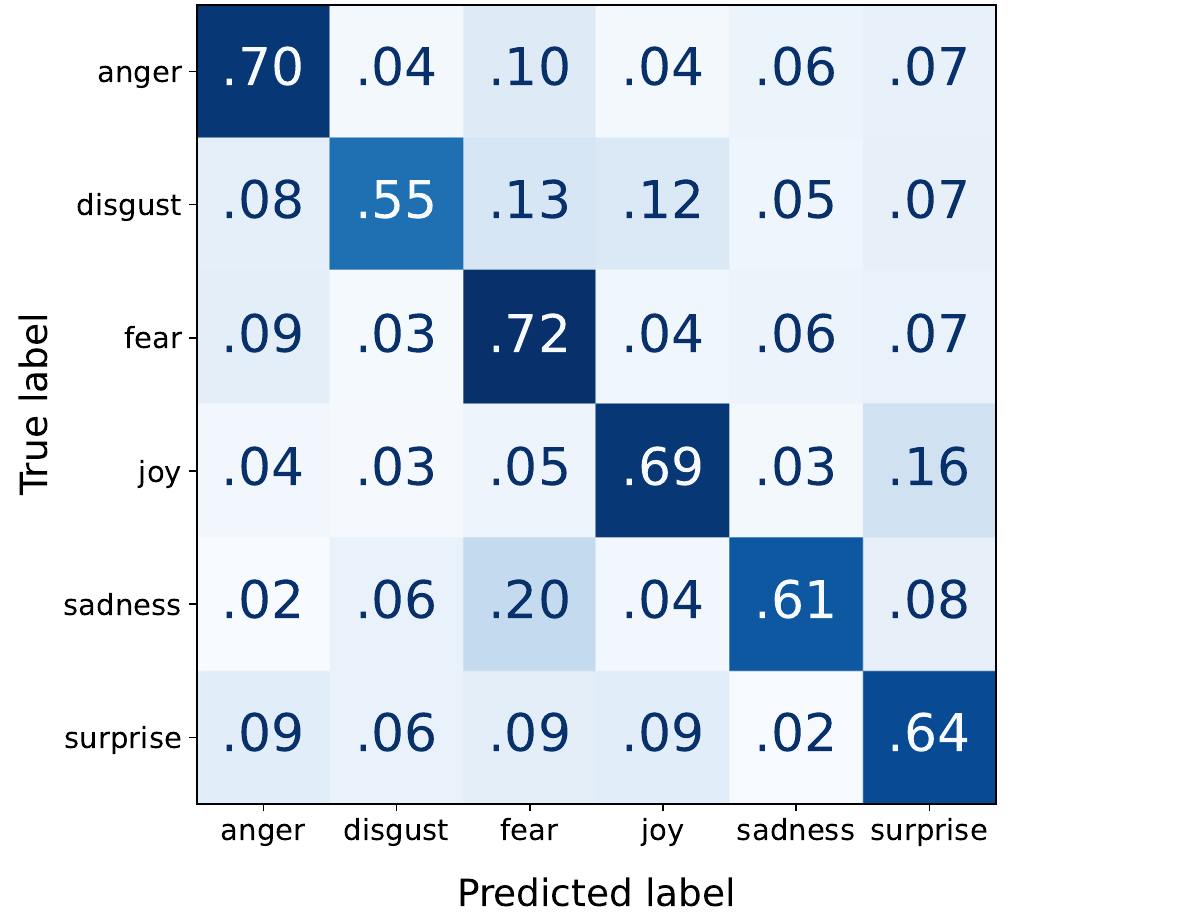}}
\caption{Confusion matrix with values normalized over true labels on the test split of Ekman-6 dataset.}
\label{fig:matrix}
\end{figure}

In Table \ref{table:results}, we present the quantitative performance of our model on the Ekman-6 dataset using the provided training and testing splits, showing that our method outperforms the state-of-the-art by $4.3\%$. Figure \ref{fig:matrix} shows the confusion matrix for our classification results on the test split.

\begin{table}[htbp]
\centering
\normalsize
\begin{tabular}{lc}
Method  & Accuracy (\%)       \\ \hline
ITE \cite{ite}     & 51.20          \\
CFN \cite{cfn}    & 51.80          \\
MART \cite{mart}   & 53.17          \\
VAANet \cite{vaanet}  & 55.30          \\
CTEN \cite{cten}   & 58.20          \\
KeyFrame \cite{keyframe} & 59.51 \\
LRCANet \cite{lrcanet} & 59.78          \\
FAEIL \cite{faeil} & 60.44          \\
TAM \cite{tam} & 61.00          \\
VEMOCLAP (Ours) & \textbf{65.28}\\
\vspace{1mm}
\end{tabular}
\label{table:results}
\caption{Classification accuracies compared to the state-of-the-art on the Ekman-6 dataset.}
\end{table}
% \vspace{-5mm}

\section{Conclusion}

In this study, we achieved a new state-of-the-art performance on the Ekman-6 video emotion classification benchmark and provided a self-contained web application for both general users and researchers. We also offer the pretrained features and highlight problematic samples from the \mbox{Ekman-6} dataset to assist researchers in improving their models. Our contributions aim to advance emotion recognition and multimedia analysis, providing valuable tools and resources to support further research and development in these fields.

\bibliographystyle{IEEEtran}
\bibliography{IEEEabrv,references}

\end{document}